\documentclass[conference]{IEEEtran}
\IEEEoverridecommandlockouts

\usepackage{amsmath,amssymb,amsfonts}
\usepackage[numbers]{natbib}
\usepackage{algorithmic}
\usepackage{graphicx}
\usepackage{textcomp}
\usepackage{xcolor}
\usepackage[ruled,vlined]{algorithm2e}
\usepackage{subcaption}
\def\BibTeX{{\rm B\kern-.05em{\sc i\kern-.025em b}\kern-.08em
    T\kern-.1667em\lower.7ex\hbox{E}\kern-.125emX}}
    
\begin{document}

\title{Deep Ensemble Learning with Frame Skipping for Face Anti-Spoofing}

\author{
    \IEEEauthorblockN{
        Usman Muhammad\IEEEauthorrefmark{1}\IEEEauthorrefmark{2}, Md Ziaul Hoque\IEEEauthorrefmark{1}, Mourad Oussalah\IEEEauthorrefmark{1} and Jorma Laaksonen \IEEEauthorrefmark{2}
    }
    \IEEEauthorblockA{\IEEEauthorrefmark{1} Center for Machine Vision and Signal Analysis, University of Oulu, Finland}
    \IEEEauthorblockA{\IEEEauthorrefmark{2} Department of Computer Science, Aalto University, Finland}
}

\maketitle

\begin{abstract}
Face presentation attacks (PA), also known as spoofing attacks, pose a substantial threat to biometric systems that rely on facial recognition systems, such as access control systems, mobile payments, and identity verification systems. To mitigate the spoofing risk, several video-based methods have been presented in the literature that analyze facial motion in successive video frames. However, estimating the motion between adjacent frames is a challenging task and requires high computational cost. In this paper, we rephrase the face anti-spoofing task as a motion prediction problem and introduce a deep ensemble learning model with a frame skipping mechanism. In particular, the proposed frame skipping adopts a uniform sampling approach by dividing the original video into video clips of fixed size. By doing so, every nth frame of the clip is selected to ensure that the temporal patterns can easily be perceived during the training of three different recurrent neural networks (RNNs). Motivated by the performance of individual RNNs, a meta-model is developed to improve the overall detection performance by combining the prediction of individual RNNs. Extensive experiments were performed on four datasets, and state-of-the-art performance is reported on  MSU-MFSD (3.12\%),  Replay-Attack (11.19\%), and OULU-NPU (12.23\%) databases by using half total error rates (HTERs) in the most challenging cross-dataset testing scenario. 
\end{abstract}


\section{Introduction}
Face recognition technology has been used in various fields, such as automated teller machines (ATMs), retail and marketing, automatic border control,  security and law enforcement \cite{muhammad2022adaptive}. However, there are various physical and digital attacks, such as 3D mask attacks \cite{liu20163d}, deep fake face swapping \cite{dong2023restricted}, face morphing \cite{ramachandra2019towards}, face adversarial attacks \cite{xu2022adversarial}, and so on, that can limit the applications of facial recognition technology. Therefore, developing robust countermeasures to detect face representation attacks is critical to improving the security of biometric systems and ensuring the widespread adoption of this technology.\\
\begin{figure}
\centering
\centerline{\includegraphics[scale=0.27]{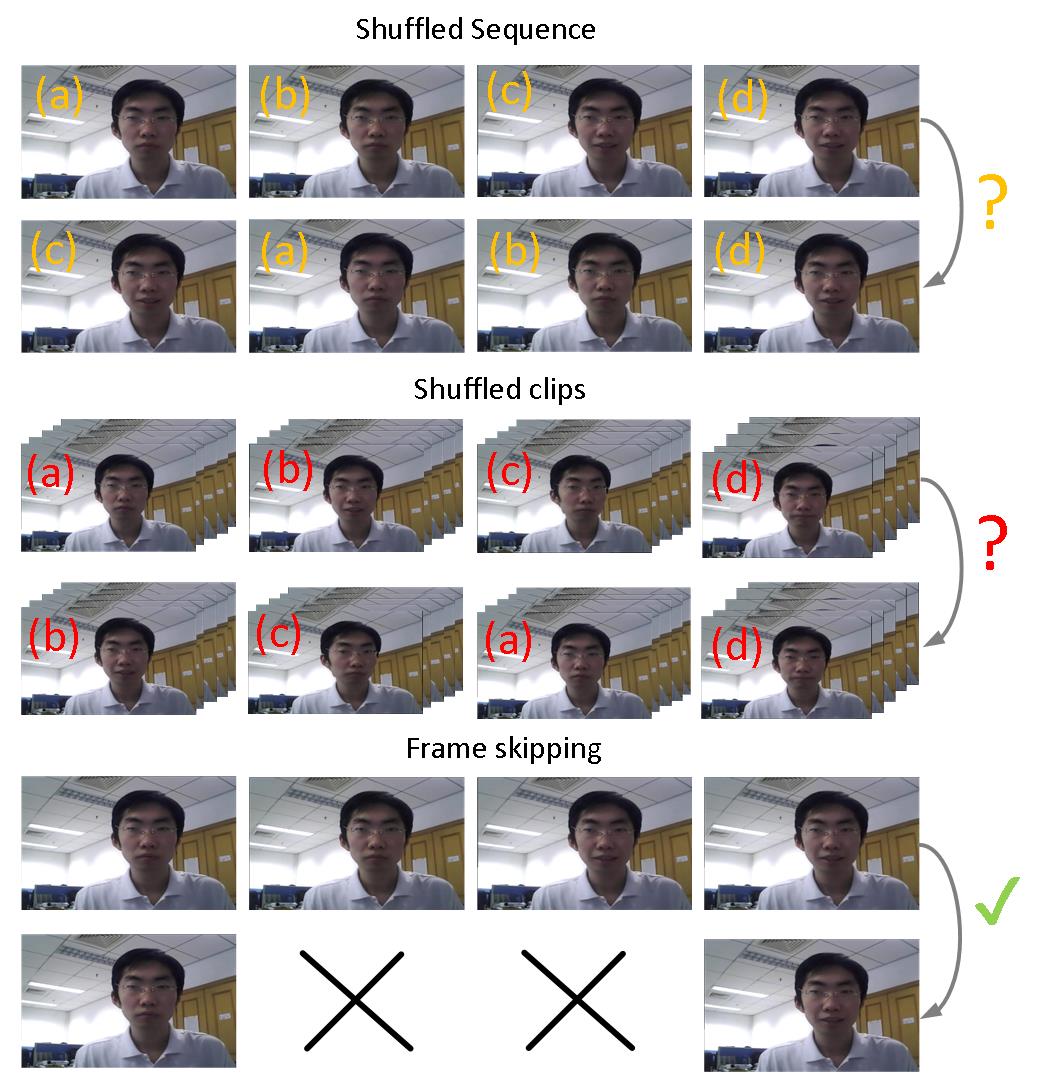}}
\vspace{0.2cm}
\caption{
Illustration of the necessity of using the frame skipping mechanism. The top four rows show sequence \cite{lee2017unsupervised} and clip prediction \cite{xu2019self}. In this work, we exploit temporal coherence in videos by proposing a frame skipping mechanism that requires a single frame in a video clip. In this way, the natural motion of the object along the time axis provides rich information to recurrent neural networks for temporal cue prediction and improves presentation attack detection.}
\label{fig.1}
\end{figure}
\begin{figure*}
\centerline{\includegraphics[scale=0.48]{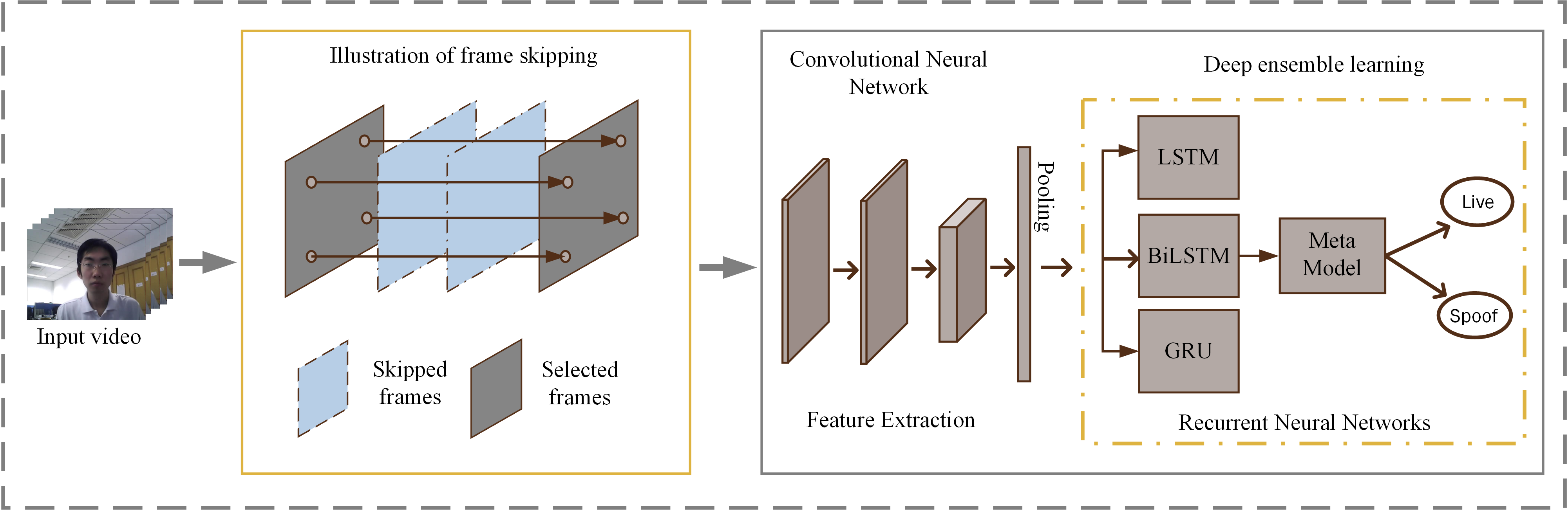}}
\vspace{0.3cm}
\caption{Flowchart of our proposed method. First, a video $X$ is divided into non-overlapping segments of smaller length $x$ to perform uniform frame sampling. Then, every nth frame of the segment is selected for training the model. As the remaining frames of the segment are skipped, we train multiple recurrent neural networks to predict the temporal cues and form a meta-model to improve generalization of PAD.}
\label{fig.2}
\end{figure*}
\indent The main issue for face PAD is to identify and analyze the unique visual and behavioral characteristics of both live and spoofed facial images because both classes contain spatiotemporal information. Recent studies show that video-based methods \cite{shao2020regularized, zhang2021two,muhammad2023self} can potentially be more effective than image-based methods \cite{boulkenafet2016face, wen2015face, muhammad2019face}. This is due to the fact that video provides additional information and context that can help discriminate real and fake faces. For instance, a real face not only exhibits the appearance of the face but also small and subtle movements, such as eye blinks and facial expressions, which can reveal important cues about the authenticity of the face \cite{muhammad2019faceb}. Moreover, temporal cues (i.e., the temporal consistency of the face over time) provide additional information, whereas a fake face may appear rigid and static. Despite the success of video-based methods, domain generalization is one of the main challenges that still need to be addressed in the face anti-spoofing domain. In our work, we define \textit{generalizability} as the extent to which a model is trained and tuned on one or multiple databases and then applied to out-of-sample unseen data.\\
\indent There are many generalization-related research topics such as ensemble learning, data augmentation, transfer learning,  meta-learning, and so on. In particular,  domain generalization is important in face anti-spoofing because new datasets are expensive and time-consuming to collect and annotate in real-world scenarios. Moreover, testing data can come from different illumination conditions or environments than the training data. Therefore, in order to improve the generalization, various methods, such as adversarial learning \cite{liu2022adversarial}, meta pattern learning \cite{cai2022learning}, generative domain adaptation \cite{zhou2022generative}, hypothesis verification \cite{liu2022feature}, or cross-adversarial learning \cite{huang2022generalized}, have been proposed that train the model on multiple datasets, but the detection performance remains limited due to a substantial distribution difference among source domains.\\
\indent In a typical PAD video, not all the frames are equally important, and processing every frame is computationally expensive \cite{muhammad2022adaptive}. Thus, frame selection approaches have been often used to reduce the computational cost. For instance, Usman {\textit{et al.}} \cite{muhammad2023face} proposed to convert the video sequences into a single RGB frame based on the Gaussian weighting function. The authors claimed that frame aggregation can amplify motion cues, such as head movements, and surface edges. Another approach to frame selection is to identify key frames that capture important moments or events in the video sequence \cite{li2018can}. In particular, the optical flow has been used for motion analysis and to identify significant changes between two adjacent frames \cite{yin2016face}. It provides a set of motion vectors that represent the motion between frames. Moreover, the global motion was found to be critical in face anti-spoofing that discriminates camera motion and natural motion of the objects along the time axis \cite{muhammad2022self}. The frame selection methods that rely on clip order prediction \cite{xu2019self} or a sequence sorting task \cite{lee2017unsupervised} in self-supervised learning enable the model to learn meaningful representations from unlabeled video data. However, these approaches extract features from all the frames of the video. We argue that a frame selection approach that involves motion estimation \cite{yin2016face, muhammad2022self, muhammad2022adaptive} between the frames can be costly to train and may lead to potential stability issues. Thus, effective handling of such spatiotemporal variations is pivotal to improving the performance of face anti-spoofing systems.\\
\indent As mentioned above, most of the motion estimation methods are designed to satisfy the PAD performance, while computational cost is mostly ignored. In an attempt to fill this gap, we propose to use a subset of frames by selecting a uniform sampling that does not require estimating the motion between adjacent frames and assumes that selected frames convey relevant information about human faces. Fig. 1 illustrates our approach. In particular, we attempt to make an alternative way of predicting the temporal changes through training different recurrent neural networks (RNNs) that model the sequential information across the video frames. Motivated by this, we also address the domain generalization issue by combining the predictions of each RNN and then develop a meta-model based on the idea of stacking-based ensemble learning \cite{muhammad2023domain}. Intuitively, this idea has at least three main advantages. First, using 4 to 7 frames is sufficient, since it does not require any additional analysis or processing of the video frames. Second, the computational cost is significantly reduced because only a few images will be enough for subsequent analysis, instead of all frames during both training and test phases. Third, we can easily control the number of frames selected from the video sequence by adjusting the sampling rate.\\
\indent The rest of this paper is organized as follows. Section II explains all the steps of the proposed method. Section III discusses the details of experimental results with state-of-the-art performance comparison. Finally, Section IV summarizes the main outcomes with conclusive statements and perspective works.
\section{Methodology}
Our work focuses on predicting the temporal dynamics across video frames. Since the 2D CNNs are well-suited for capturing spatial information (i.e., edges, corners, and textures), we use different variants of recurrent neural networks for capturing temporal dependencies across video frames. Firstly, frame skipping is applied to select only a subset of the frames, which can help to reduce the need of computational resources and improve the efficiency of the analysis. Secondly, a CNN is utilized to extract spatial information. Then, several RNN-based models are trained on the same dataset, and their predictions are combined into a single meta-model. Each of the above steps is explained in the following sub-sections.

\subsection{Frame Skipping Mechanism}
Selecting important frames for face anti-spoofing is a relatively new area. A fundamental yet challenging problem in detecting real and attack videos is dealing with temporal variations. Since there is no fixed duration of real and attack videos in most of the PAD datasets, this boils down to the question of how many frames in the video contribute to face anti-spoofing. Moreover, the high computational cost associated with the optical flow method makes it impractical for real-time applications or large-scale video processing tasks.
Conventionally, motion estimation is performed between every two adjacent frames in a video clip. However, in the context of face anti-spoofing, the frames that show changes in facial expression or movement between two adjacent frames also vary. Thus, computing the motion between every pair of frames in a video clip can be computationally expensive, especially when two adjacent frames are highly similar. Motivated by the aforementioned observations and the emergence of new methods for focusing on the most relevant frames only \cite{muhammad2023face}, we propose the integration of a frame skipping mechanism to show that 4 to 7 frames are sufficient for state-of-the-art face presentation attack detection. In particular, we adopt a uniform frame sampling method that selects a subset of frames from a video sequence to improve the efficiency of the detection process. More specifically, to generate a subset of frames, a video is equally divided into non-overlapping segments where the total number of frames in the video are represented as $T$, and each segment consists of thirty frames. Let's assume that $n$ segments are generated, and only the last frame of each segment is saved. Thus, we can use the following equation:
\begin{equation}
L = min(T, n \cdot 30)
\end{equation}
where $L$ represents the last frame saved, $T$ is the total number of frames in the video, $n$ is the number of segments, and we take the minimum between $T$ and $n \cdot 30$ to ensure that we only select and save one frame from each segment while considering the available frames in the video.

\begin{algorithm}[]
  \SetAlgoLined
  \KwIn{ $I=\left \{ a_{i},b_{i} \right \}_{j=1}^{l}$}
  \KwOut{$R_{f}$}

  \For{$i \leftarrow 1$ \KwTo $N$}{
    Learn $L_1$ based on $I$\hspace{0.5cm} \tcp{$L_1$ is LSTM with its hyper parameter}
  }

  \For{$i \leftarrow 1$ \KwTo $N$}{
    Learn $L_2$ based on $I$\hspace{0.5cm} \tcp{$L_2$ is BiLSTM  with its hyper parameter}
  }

    \For{$i \leftarrow 1$ \KwTo $N$}{
    Learn $L_3$ based on $I$\hspace{0.5cm} \tcp{$L_3$ is GRU with its hyper parameter}
  }

  \For{$i \leftarrow 1$ \KwTo $N$}{ \ $R_{1}=I(L_{1})$ \hspace{0.5cm} \tcp{Predict $R_1$ given data $I$ and model $L_{1}$}\ 
   $R_{2}=I(L_{2})$ \hspace{0.5cm} \tcp{Predict $R_2$ given data $I$ and model $L_{2}$}\ 
   $R_{3}=I(L_{3})$ \hspace{0.5cm} \tcp{Predict $R_3$ given data $I$ and model $L_{3}$}\ 
  }
  

   \For{$n \leftarrow 1$ \KwTo $l$}{
   $R_{f} = (R_{1}  +R_{2} + R_{3})$}
\caption{Deep Ensemble Learning}
\end{algorithm}

\setlength{\tabcolsep}{8pt}
\begin{table*}
\begin{center}
\caption{Performance evaluation using MSU-MFSD (M), CASIA-MFSD (C), Replay-Attack (I), and OULU-NPU (0) databases. Comparison results are obtained directly from the corresponding papers in terms of HTER(\%) and AUC(\%).} \label{tab:cap3}
\resizebox{13cm}{!}{%

\begin{tabular}{l|r|l|r | l| l | l|l | l}
\hline
 & \multicolumn{2}{c|}{O\&C\&I to M} & \multicolumn{2}{c|}{O\&M\&I to C}  & \multicolumn{2}{c|}{O\&C\&M to I} & \multicolumn{2}{c}{I\&C\&M to O} \\ \hline
  Method     & HTER   & AUC   & HTER   & AUC  & HTER    & AUC    & HTER  & AUC   \\ \hline 
MADDG  \cite{shao2019multi} & 17.69    & 88.06   &  24.50 &  84.51 & 22.19 &   84.99 &  27.89 &   80.02 \\
DAFL  \cite{saha2020domain} & 14.58    & 92.58   &  17.41 &  90.12 & 15.13 &  95.76 &  14.72 &  93.08 \\ 
SSDG-R  \cite{jia2020single} & 7.38    & 97.17   &  10.44 &  95.94 & \underline{11.71} &   \textbf{96.59} &  15.61 &   91.54 \\ 
DR-MD  \cite{wang2020cross} & 17.02   & 90.10 &  19.68 &  87.43 & 20.87 & 86.72 &  25.02 &  81.47 \\
MA-Net  \cite{liu2021face} & 20.80    &  ---  &  25.60 & ---  & 24.70 & ---  &  26.30 &  ---  \\
RFMetaFAS  \cite{shao2020regularized} & 13.89    &  93.98  &  20.27 &  88.16 & 17.30 & 90.48 &  16.45 &  91.16 \\
FAS-DR-BC(MT)  \cite{qin2021meta} & 11.67    & 93.09   &  18.44 &  89.67 & 11.93 & \underline{94.95} &  16.23 &  91.18 \\
ASGS \cite{muhammad2022adaptive}  & 5.91   & \underline{99.88} &   10.21  & \textbf{99.86}  &  45.84 &  76.09  &  13.54  &   \textbf{99.73}  \\
HFN + MP  \cite{cai2022learning} & 5.24 & 97.28 &  \textbf{9.11}  &   96.09 &   15.35 &   90.67  & \underline{12.40} &  94.26 \\ 
Cross-ADD \cite{huang2022generalized}  & 11.64   & 95.27 &   17.51  & 89.98  &  15.08 &  91.92  &  14.27  &    93.04  \\ 
FG +HV  \cite{liu2022feature} & 9.17   & 96.92  &  12.47 &  93.47 & 16.29  &  90.11 &  13.58 &  93.55 \\
ADL  \cite{liu2022adversarial} &   5.00   &   97.58    & 10.00 &  96.85 & 12.07 &   94.68  & 13.45  &  94.43 \\ 
           \hline 
CNN-LSTM & 3.65   &  99.71   &  18.29   &  \underline{99.81}    & 13.99    &   88.62   &  16.00   &  87.36 \\ 
CNN-BiLSTM  & 5.31 &  99.40  &  20.74 &  90.53 & 12.01 &  90.50 &  22.88 &  85.33 \\
CNN-GRU  & 4.74 &  99.02  &  36.41 &  99.34 & 17.41 & 87.18 &  17.94 &  94.93 \\
 Meta-Model (Ours) &  \textbf{3.12} &  \textbf{99.89}  & 23.71  & 99.66 &  \textbf{11.19}  & 90.95  &  \textbf{12.23} & \underline{97.70} \\ 
\hline
\end{tabular}}
\end{center}
\end{table*}

\begin{figure*}
   \centering
       \begin{subfigure}[b]{0.23\textwidth}
        \centering
          \includegraphics[width=\textwidth]{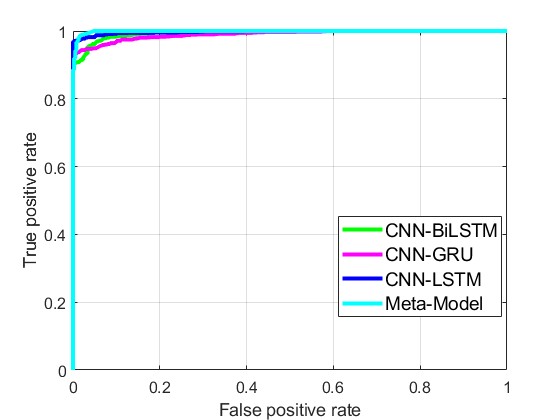}
         \caption{}
     \end{subfigure}
     \hfill
     \begin{subfigure}[b]{0.23\textwidth}
        \centering
          \includegraphics[width=\textwidth]{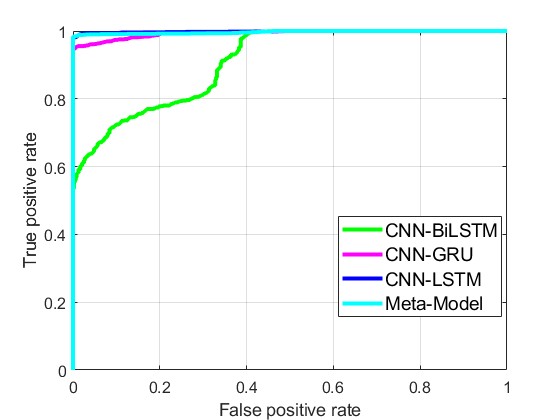}
         \caption{}
     \end{subfigure}
     \hfill
     \begin{subfigure}[b]{0.23\textwidth}
        \centering
         \includegraphics[width=\textwidth]{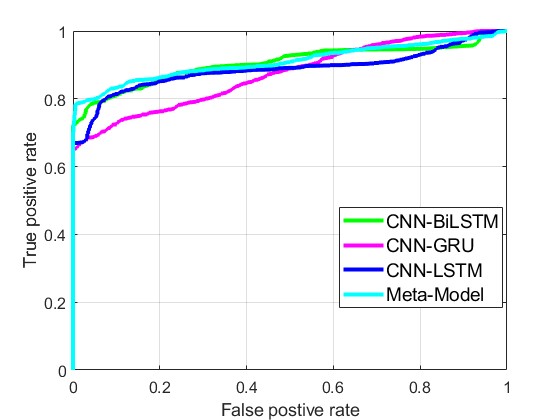}
         \caption{}
     \end{subfigure}
      \hfill
     \begin{subfigure}[b]{0.23\textwidth}
        \centering
          \includegraphics[width=\textwidth]{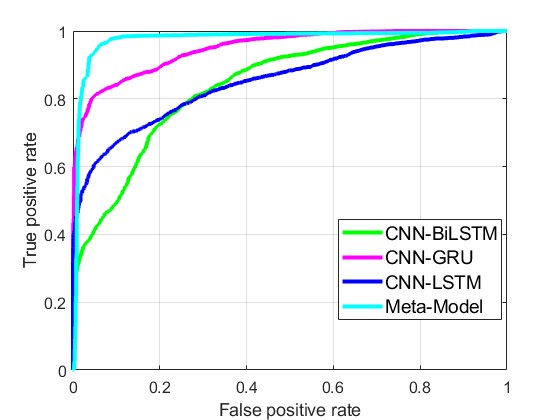}
         \caption{}
     \end{subfigure}
          \hfill
        \caption{The Receiver Operating Characteristics (ROC) curves. (a) O\&C\&I to M, (b) O\&M\&I to C, (c) O\&C\&M to I, and (d) I\&C\&M to O illustrates ROC curves for four datasets.\label{fig.3}}
\end{figure*}

\subsection{Deep Ensemble Learning}
An illustration of the overall framework is provided in Fig. 2. First, we employed a pre-trained CNN for extracting spatial information. Our proposed feature extraction part extracts the deep features from the pooling layer of a CNN model. In the next stage, instead of training several different CNN architectures, three different RNNs are trained to get the benefit of the frame skipping mechanism. To achieve this, ensemble learning is proposed by combining the strength of each RNN model to form a single meta-model. The meta-model takes the predictions from the base models as input and produces a final detection. Our motivation behind stacking is that by combining the predictions of multiple models, the overall performance of the system can be improved. Stacking, also known as a stacked generalization, is a technique used in ensemble learning, which involves training a meta-model that learns to combine the predictions of multiple base models \cite{muhammad2023domain}. In our work, three bases models, such as Long short-term memory (LSTM) \cite{hochreiter1997long}, Bidirectional long short-term memory (BiLSTM) \cite{schuster1997bidirectional}, and Gated recurrent unit (GRU) \cite{cho2014learning} are adopted.\\ 
\indent The first base model (LSTM) captures the temporal patterns and dynamics across frames by processing the CNN features sequentially. In particular, LSTM analyzes the sequential input, one by one, maintaining an internal memory that retains information from previous sequence while considering the current sequence. Specifically, LSTM contains memory cells, which enable it to selectively remember or forget information over time  \cite{hochreiter1997long}. The second base model (BiLSTM) processes the sequential features in temporal order, either forward or backward,  allowing it to capture both past and future temporal dependencies in the video data \cite{schuster1997bidirectional}. The third base model (GRU) showed that gating is indeed helpful in general and useful to capture the temporal dependencies of the face over time \cite{cho2014learning}. The model processes visual features in a temporal sequence by updating its hidden state based on the current input and its previous hidden state.\\ 
\indent After training these sub-models, we simply concatenate the outcome (predictions) of all the models and then train another model based on LSTM \cite{hochreiter1997long} architecture. We call this meta-model because it is applied to leverage the strengths of individual base models by integrating their predictions in a way that complements the overall predictive power. Pseudo-code is provided in  Algorithm 1.

\section{Experimental Setup}
To evaluate the performance of the proposed ensemble learning, OULU-NPU database \cite{boulkenafet2017oulu} (denoted as O), MSU Mobile Face Spoofing database \cite{wen2015face} (denoted as M), Idiap Replay-Attack database \cite{chingovska2012effectiveness} (denoted as I), and CASIA Face Anti-Spoofing database (denoted as C) are used in our work. The Half Total Error Rate (HTER) is reported by taking the average of the false acceptance rate (FAR) and the false rejection rate (FRR).

\subsection{Implementation details}
The Densely Connected Convolutional Networks (DenseNet-201) \cite{huang2017densely} architecture is employed and all the frames are resized to  $224 \times 224$ to extract deep features. In particular, $1920$-dimensional features from the last pooling layer are extracted without fine-tuning the model. For the ensemble learning, the LSTM is trained by using the Adam optimizer with the hidden layer dimension of $1000$, validation frequency $30$, and mini-batch size of $32$. The He initializer \cite{he2015delving} is utilized as weight initialization technique and the learning rate is adjusted to $0.0001$. Since no fixed size epochs were used, an early stopping function \cite{prechelt2002early} was utilized to prevent overfitting and improve the generalization of the model. For training the BiLSTM model, we keep the same parameters as used for training the LSTM except for the size of the hidden layer dimension which is set to $500$. Finally, the third model, GRU, also follows the same tuning parameters and only the hidden layer size was decreased to $20$. To develop our meta-model, we train the LSTM model based on the hidden layer dimension $20$, epochs size (i.e., $100$), and by following the same parameters used to train the sub-models.

\subsection{Comparison against the state-of-the-art methods}
In Table I, we report the performance of the proposed method against state-of-the-art deep learning methods. In particular, various domain generalization methods train the model on three source databases and test it on a completely unseen database using the leave-one-out (LOO) strategy. To make a fair comparison with them, we first compute the equal error rate (EER) on the testing set of source databases and then HTER is calculated directly on the target (unseen) dataset. One can see that the proposed meta-model achieves the best performance on three protocols of O\&C\&I to M, O\&C\&M to I, and I\&C\&M to O. We also use another evaluation metric such as area under the ROC Curve (AUC) to measures the overall performance of a classifier by measuring the trade-off between the true positive rate (sensitivity) and false positive rate (1 -- specificity). It can be observed that deep ensemble learning achieves more than $90\%$ AUC on all the datasets. In comparison to AUC, HTER shows decreased performance than AUC because HTER focuses on the equal error rate threshold and may not be the most relevant or representative measure of overall performance. Moreover, we also visualize ROC curves in Fig. 3. The meta-model (ensemble learning) indicates a better discrimination capability with a higher curve (closer to the upper left corner).

In comparison to other domain generalization-related methods, such as adversarial learning \cite{liu2022adversarial}, disentangled representation learning \cite{wang2020cross}, and meta-learning \cite{shao2020regularized, cai2022learning, qin2021meta}, the proposed ensemble learning demonstrates that predicting the motion across video frames can lead to a better generalization capability. Furthermore, since the motion is not computed between adjacent frames, we do not compare the computational time with any other motion estimation method such as optical flow. Thus, the proposed skipping mechanism reduces the amount of data to process, allowing for faster analysis and only sacrificing the performance on one dataset (i.e., CASIA) in comparison to state-of-the-art methods.

\section{Conclusions}
In this paper, we addressed the domain generalization issue based on ensemble learning and the frame skipping mechanism. In particular, the skipping mechanism provides an alternative way of predicting the temporal changes through training different recurrent neural networks rather than proposing a motion estimation method between the adjacent frames. Since only certain frames were involved during training, this approach reduces the computational load and can be implemented in a scenario where real-time performance is crucial. Moreover, we show that the performance of LSTM, BiLSTM and GRU remains limited without using the meta-model. Thus, we conclude that the proposed ensemble learning can compensate the weaknesses of multiple sub-models and reduce the overall error rate.  Based on the experimental results on four benchmark datasets, the proposed method exhibits state-of-the-art performance on three datasets. However, our approach may not be suitable for applications that require precise frame-by-frame analysis or rely heavily on temporal information. Thus, our future work will focus on the development of deep learning-based methods that can estimate the motion directly between the frames in end-to-end learning.

\section{Declaration of Competing Interest}
The authors have no conflict of interest that could have appeared to influence the work reported in this paper.

\section*{Acknowledgment}
This work is financially supported by ‘Understanding speech and scene with ears and eyes (USSEE)’’ (project number 345791). The first author also acknowledges the support of the Ella and Georg Ehrnrooth foundation.

\bibliographystyle{IEEEtran}
\bibliography{strings.bib}

\begin{thebibliography}{10}
\providecommand{\url}[1]{#1}
\csname url@samestyle\endcsname
\providecommand{\newblock}{\relax}
\providecommand{\bibinfo}[2]{#2}
\providecommand{\BIBentrySTDinterwordspacing}{\spaceskip=0pt\relax}
\providecommand{\BIBentryALTinterwordstretchfactor}{4}
\providecommand{\BIBentryALTinterwordspacing}{\spaceskip=\fontdimen2\font plus
\BIBentryALTinterwordstretchfactor\fontdimen3\font minus
  \fontdimen4\font\relax}
\providecommand{\BIBforeignlanguage}[2]{{%
\expandafter\ifx\csname l@#1\endcsname\relax
\typeout{** WARNING: IEEEtran.bst: No hyphenation pattern has been}%
\typeout{** loaded for the language `#1'. Using the pattern for}%
\typeout{** the default language instead.}%
\else
\language=\csname l@#1\endcsname
\fi
#2}}
\providecommand{\BIBdecl}{\relax}
\BIBdecl

\bibitem{muhammad2022adaptive}
U.~Muhammad, J.~Zhang, L.~Liu, and M.~Oussalah, ``An adaptive spatio-temporal
  global sampling for presentation attack detection,'' \emph{IEEE Transactions
  on Circuits and Systems II: Express Briefs}, 2022.

\bibitem{liu20163d}
S.~Liu, B.~Yang, P.~C. Yuen, and G.~Zhao, ``A 3d mask face anti-spoofing
  database with real world variations,'' in \emph{Proceedings of the IEEE
  conference on computer vision and pattern recognition workshops}, 2016, pp.
  100--106.

\bibitem{dong2023restricted}
J.~Dong, Y.~Wang, J.~Lai, and X.~Xie, ``Restricted black-box adversarial attack
  against deepfake face swapping,'' \emph{IEEE Transactions on Information
  Forensics and Security}, 2023.

\bibitem{ramachandra2019towards}
R.~Ramachandra, S.~Venkatesh, K.~Raja, and C.~Busch, ``Towards making morphing
  attack detection robust using hybrid scale-space colour texture features,''
  in \emph{2019 IEEE 5th International Conference on Identity, Security, and
  Behavior Analysis (ISBA)}.\hskip 1em plus 0.5em minus 0.4em\relax IEEE, 2019,
  pp. 1--8.

\bibitem{xu2022adversarial}
Y.~Xu, K.~Raja, R.~Ramachandra, and C.~Busch, ``Adversarial attacks on face
  recognition systems,'' in \emph{Handbook of Digital Face Manipulation and
  Detection: From DeepFakes to Morphing Attacks}.\hskip 1em plus 0.5em minus
  0.4em\relax Springer International Publishing Cham, 2022, pp. 139--161.

\bibitem{lee2017unsupervised}
H.-Y. Lee, J.-B. Huang, M.~Singh, and M.-H. Yang, ``Unsupervised representation
  learning by sorting sequences,'' in \emph{Proceedings of the IEEE
  international conference on computer vision}, 2017, pp. 667--676.

\bibitem{xu2019self}
D.~Xu, J.~Xiao, Z.~Zhao, J.~Shao, D.~Xie, and Y.~Zhuang, ``Self-supervised
  spatiotemporal learning via video clip order prediction,'' in
  \emph{Proceedings of the IEEE/CVF Conference on Computer Vision and Pattern
  Recognition}, 2019, pp. 10\,334--10\,343.

\bibitem{shao2020regularized}
R.~Shao, X.~Lan, and P.~C. Yuen, ``Regularized fine-grained meta face
  anti-spoofing,'' in \emph{Proceedings of the AAAI Conference on Artificial
  Intelligence}, 2020, pp. 11\,974--11\,981.

\bibitem{zhang2021two}
Z.~Zhang, C.~Jiang, X.~Zhong, C.~Song, and Y.~Zhang, ``Two-stream convolutional
  networks for multi-frame face anti-spoofing,'' \emph{arXiv preprint
  arXiv:2108.04032}, 2021.

\bibitem{muhammad2023self}
U.~Muhammad and M.~Oussalah, ``Self-supervised face presentation attack
  detection with dynamic grayscale snippets,'' in \emph{2023 IEEE 17th
  International Conference on Automatic Face and Gesture Recognition
  (FG)}.\hskip 1em plus 0.5em minus 0.4em\relax IEEE, 2023, pp. 1--6.

\bibitem{boulkenafet2016face}
Z.~Boulkenafet, J.~Komulainen, and A.~Hadid, ``Face spoofing detection using
  colour texture analysis,'' \emph{IEEE Transactions on Information Forensics
  and Security}, vol.~11, no.~8, pp. 1818--1830, 2016.

\bibitem{wen2015face}
D.~Wen, H.~Han, and A.~K. Jain, ``Face spoof detection with image distortion
  analysis,'' \emph{IEEE Transactions on Information Forensics and Security},
  vol.~10, no.~4, pp. 746--761, 2015.

\bibitem{muhammad2019face}
U.~Muhammad and A.~Hadid, ``Face anti-spoofing using hybrid residual learning
  framework,'' in \emph{2019 International Conference on Biometrics
  (ICB)}.\hskip 1em plus 0.5em minus 0.4em\relax IEEE, 2019, pp. 1--7.

\bibitem{muhammad2019faceb}
U.~Muhammad, T.~Holmberg, W.~Carneiro~de Melo, and A.~Hadid, ``Face
  anti-spoofing via sample learning based recurrent neural network (rnn),'' in
  \emph{The British Machine Vision Conference 2019 (BMVC) 9th-12th September
  2019, Cardiff UK}.\hskip 1em plus 0.5em minus 0.4em\relax British Machine
  Vision Association Press, 2019.

\bibitem{liu2022adversarial}
M.~Liu, J.~Mu, Z.~Yu, K.~Ruan, B.~Shu, and J.~Yang, ``Adversarial learning and
  decomposition-based domain generalization for face anti-spoofing,''
  \emph{Pattern Recognition Letters}, vol. 155, pp. 171--177, 2022.

\bibitem{cai2022learning}
R.~Cai, Z.~Li, R.~Wan, H.~Li, Y.~Hu, and A.~C. Kot, ``Learning meta pattern for
  face anti-spoofing,'' \emph{IEEE Transactions on Information Forensics and
  Security}, vol.~17, pp. 1201--1213, 2022.

\bibitem{zhou2022generative}
Q.~Zhou, K.-Y. Zhang, T.~Yao, R.~Yi, K.~Sheng, S.~Ding, and L.~Ma, ``Generative
  domain adaptation for face anti-spoofing,'' in \emph{Computer Vision--ECCV
  2022: 17th European Conference, Tel Aviv, Israel, October 23--27, 2022,
  Proceedings, Part V}.\hskip 1em plus 0.5em minus 0.4em\relax Springer, 2022,
  pp. 335--356.

\bibitem{liu2022feature}
S.~Liu, S.~Lu, H.~Xu, J.~Yang, S.~Ding, and L.~Ma, ``Feature generation and
  hypothesis verification for reliable face anti-spoofing,'' in
  \emph{Proceedings of the AAAI Conference on Artificial Intelligence}, 2022,
  pp. 1782--1791.

\bibitem{huang2022generalized}
H.~Huang, Y.~Xiang, G.~Yang, L.~Lv, X.~Li, Z.~Weng, and Y.~Fu, ``Generalized
  face anti-spoofing via cross-adversarial disentanglement with mixing
  augmentation,'' in \emph{ICASSP 2022-2022 IEEE International Conference on
  Acoustics, Speech and Signal Processing (ICASSP)}.\hskip 1em plus 0.5em minus
  0.4em\relax IEEE, 2022, pp. 2939--2943.

\bibitem{muhammad2023face}
U.~Muhammad and M.~Oussalah, ``Face anti-spoofing from the perspective of data
  sampling,'' \emph{Electronics Letters}, vol.~59, no.~1, p. e12692, 2023.

\bibitem{li2018can}
Y.~Li, X.~Huang, and G.~Zhao, ``Can micro-expression be recognized based on
  single apex frame?'' in \emph{2018 25th IEEE International Conference on
  Image Processing (ICIP)}.\hskip 1em plus 0.5em minus 0.4em\relax IEEE, 2018,
  pp. 3094--3098.

\bibitem{yin2016face}
W.~Yin, Y.~Ming, and L.~Tian, ``A face anti-spoofing method based on optical
  flow field,'' in \emph{2016 IEEE 13th International Conference on Signal
  Processing (ICSP)}.\hskip 1em plus 0.5em minus 0.4em\relax IEEE, 2016, pp.
  1333--1337.

\bibitem{muhammad2022self}
U.~Muhammad, Z.~Yu, and J.~Komulainen, ``Self-supervised {2D} face presentation
  attack detection via temporal sequence sampling,'' \emph{Pattern Recognition
  Letters}, vol. 156, pp. 15--22, 2022.

\bibitem{muhammad2023domain}
U.~Muhammad, D.~R. Beddiar, and M.~Oussalah, ``Domain generalization via
  ensemble stacking for face presentation attack detection,'' \emph{arXiv
  preprint arXiv:2301.02145}, 2023.

\bibitem{shao2019multi}
R.~Shao, X.~Lan, J.~Li, and P.~C. Yuen, ``Multi-adversarial discriminative deep
  domain generalization for face presentation attack detection,'' in
  \emph{Proceedings of the IEEE/CVF conference on computer vision and pattern
  recognition}, 2019, pp. 10\,023--10\,031.

\bibitem{saha2020domain}
S.~Saha, W.~Xu, M.~Kanakis, S.~Georgoulis, Y.~Chen, D.~P. Paudel, and
  L.~Van~Gool, ``Domain agnostic feature learning for image and video based
  face anti-spoofing,'' in \emph{Proceedings of the IEEE/CVF Conference on
  Computer Vision and Pattern Recognition Workshops}, 2020, pp. 802--803.

\bibitem{jia2020single}
Y.~Jia, J.~Zhang, S.~Shan, and X.~Chen, ``Single-side domain generalization for
  face anti-spoofing,'' in \emph{Proceedings of the IEEE/CVF Conference on
  Computer Vision and Pattern Recognition}, 2020, pp. 8484--8493.

\bibitem{wang2020cross}
G.~Wang, H.~Han, S.~Shan, and X.~Chen, ``Cross-domain face presentation attack
  detection via multi-domain disentangled representation learning,'' in
  \emph{Proceedings of the IEEE/CVF Conference on Computer Vision and Pattern
  Recognition}, 2020, pp. 6678--6687.

\bibitem{liu2021face}
A.~Liu, Z.~Tan, J.~Wan, Y.~Liang, Z.~Lei, G.~Guo, and S.~Z. Li, ``Face
  anti-spoofing via adversarial cross-modality translation,'' \emph{IEEE
  Transactions on Information Forensics and Security}, vol.~16, pp. 2759--2772,
  2021.

\bibitem{qin2021meta}
Y.~Qin, Z.~Yu, L.~Yan, Z.~Wang, C.~Zhao, and Z.~Lei, ``Meta-teacher for face
  anti-spoofing,'' \emph{IEEE transactions on pattern analysis and machine
  intelligence}, vol.~44, no.~10, pp. 6311--6326, 2021.

\bibitem{hochreiter1997long}
S.~Hochreiter and J.~Schmidhuber, ``Long short-term memory,'' \emph{Neural
  computation}, vol.~9, no.~8, pp. 1735--1780, 1997.

\bibitem{schuster1997bidirectional}
M.~Schuster and K.~K. Paliwal, ``Bidirectional recurrent neural networks,''
  \emph{IEEE transactions on Signal Processing}, vol.~45, no.~11, pp.
  2673--2681, 1997.

\bibitem{cho2014learning}
K.~Cho, B.~Van~Merri{\"e}nboer, C.~Gulcehre, D.~Bahdanau, F.~Bougares,
  H.~Schwenk, and Y.~Bengio, ``Learning phrase representations using {RNN}
  encoder-decoder for statistical machine translation,'' \emph{arXiv preprint
  arXiv:1406.1078}, 2014.

\bibitem{boulkenafet2017oulu}
Z.~Boulkenafet, J.~Komulainen, L.~Li, X.~Feng, and A.~Hadid, ``{OULU-NPU}: A
  mobile face presentation attack database with real-world variations,'' in
  \emph{2017 12th IEEE international conference on automatic face \& gesture
  recognition (FG 2017)}.\hskip 1em plus 0.5em minus 0.4em\relax IEEE, 2017,
  pp. 612--618.

\bibitem{chingovska2012effectiveness}
I.~Chingovska, A.~Anjos, and S.~Marcel, ``On the effectiveness of local binary
  patterns in face anti-spoofing,'' in \emph{2012 BIOSIG-proceedings of the
  international conference of biometrics special interest group
  (BIOSIG)}.\hskip 1em plus 0.5em minus 0.4em\relax IEEE, 2012, pp. 1--7.

\bibitem{huang2017densely}
G.~Huang, Z.~Liu, L.~Van Der~Maaten, and K.~Q. Weinberger, ``Densely connected
  convolutional networks,'' in \emph{Proceedings of the IEEE conference on
  computer vision and pattern recognition}, 2017, pp. 4700--4708.

\bibitem{he2015delving}
K.~He, X.~Zhang, S.~Ren, and J.~Sun, ``Delving deep into rectifiers: Surpassing
  human-level performance on {ImageNet} classification,'' in \emph{Proceedings
  of the IEEE international conference on computer vision}, 2015, pp.
  1026--1034.

\bibitem{prechelt2002early}
L.~Prechelt, ``Early stopping-but when?'' in \emph{Neural Networks: Tricks of
  the trade}.\hskip 1em plus 0.5em minus 0.4em\relax Springer, 2002, pp.
  55--69.

\end{thebibliography}

\end{document}